\documentclass[10pt, a4paper]{article}

\usepackage{lrec2026}

\usepackage{stfloats}      
\usepackage{amsmath}
\usepackage{float}
\usepackage{amssymb}
\usepackage{amsfonts}
\usepackage[T1]{fontenc}
\usepackage{caption}
\usepackage[utf8]{inputenc}
\usepackage{booktabs}
\usepackage{siunitx}
\usepackage{microtype}
\usepackage{inconsolata}
\usepackage{multirow}
\usepackage{graphicx}
\usepackage{color}
\usepackage{CJK}

\title{CNSocialDepress: A Chinese Social Media Dataset for Depression Risk Detection and Structured Analysis}

\name{%
\shortstack{%
\bfseries
Jinyuan XU\textsuperscript{1*},
Tian LAN\textsuperscript{2*},
Xintao YU\textsuperscript{3},
Xue HE\textsuperscript{3,4},
Hezhi ZHANG\textsuperscript{5},
Ying WANG\textsuperscript{6}\\
Pierre Magistry\textsuperscript{1},
Mathieu Valette\textsuperscript{1},
Lei LI\textsuperscript{7\dag}
}}

\address{\textsuperscript{1}Ertim Inalco, \textsuperscript{2}Milkuya Studio, \textsuperscript{3}Sorbonne Université, \textsuperscript{4}IRD Lab, \\
         \textsuperscript{5}Faculty of Psychology, Peking University, \\
         \textsuperscript{6}Faculty of Psychology and Cognitive Science, Beijing Normal University, \\
         \textsuperscript{7}Beijing Institute of Technology
        }

\abstract{
Depression is a pressing global public health issue, yet publicly available Chinese-language resources for depression risk detection remain scarce and largely focus on binary classification. To address this limitation, we release \href{https://github.com/jytal/CNSocialDepress}{\textbf{CNSocialDepress}}, a benchmark dataset for depression risk detection on Chinese social media. The dataset contains 44,178 posts from 233 users; psychological experts annotated 10,306 depression-related segments. CNSocialDepress provides binary risk labels along with structured, multidimensional psychological attributes, enabling interpretable and fine-grained analyses of depressive signals. Experimental results demonstrate the dataset's utility across a range of NLP tasks, including structured psychological profiling and fine-tuning large language models for depression detection. Comprehensive evaluations highlight the dataset's effectiveness and practical value for depression risk identification and psychological analysis, thereby providing insights for mental health applications tailored to Chinese-speaking populations.
\\ \newline }

\begin{document}

\maketitleabstract
\begingroup
\renewcommand\thefootnote{\fnsymbol{footnote}}
\footnotetext[1]{These authors contributed equally to this work.}
\footnotetext[2]{Corresponding Author.}
\endgroup

\section{Introduction}

Depressive disorders are among the most common mental health conditions worldwide and are characterized by persistent low mood or loss of interest in daily activities. According to a 2023 report\footnote{\url{https://www.who.int/news-room/fact-sheets/detail/depression}} by the World Health Organization (WHO), approximately 280 million people worldwide live with depression. In China, a 2024 survey by the Chinese Center for Disease Control and Prevention \cite{wang2024association} estimates that around 95 million individuals are affected by depression. Of the approximately 280{,}000 suicides reported annually in China, 40\% are linked to depressive disorders. Moreover, research \cite{wang2024association} shows that depression is strongly associated with both suicidal behavior and non-suicidal self-injury.

Motivated by the urgent need to detect depression, researchers increasingly apply machine learning (ML) and natural language processing (NLP) methods to automatically assess depression risk \cite{squires2023deep,hasib2023depression,aleem2022machine,liu2024graph,shi2024scaling,jia2024adaptive}. While early efforts have shown promising results, they remain constrained by the limitations of existing datasets. Traditional depression detection studies often rely on clinical data~\cite{bittar2019text,fernandes2018identifying} or on transcripts from medical or psychological interviews~\cite{shen2022automatic,li2022bidirectional}, which are expensive to collect, limited in size and diversity, and may not reflect the informal, emotionally nuanced expressions typical of real-world online environments.

To overcome these limitations, recent research has shifted toward leveraging user-generated content on social media platforms \cite{bucur2025datasets,harrigian-etal-2021-state,li2025human,jiang2024back,cai2025bayesian}, which provides rich linguistic signals and is more readily available. Such data are also more easily anonymized, thereby alleviating some privacy concerns. Beyond these methodological advantages, social media has become an important channel for emotional expression, especially in East Asian contexts \cite{zhou2023feeling,yang2009intergenerational,zhou2024infant,yang2025you}, where personal emotions may be expressed more freely online than in face-to-face settings. As a result, social media can provide richer and more authentic linguistic signals for depression analysis.

However, most existing datasets focus solely on classification with binary or multi-class labels and lack structured psychological insights or professional validation. Moreover, generative models are increasingly used in mental health applications~\cite{hu2024psycollmenhancingllmpsychological,xu2024mental,yang2024mentallama,lai2023psy,He2025Modular,gu2025mocount,cai2025role, xu-etal-2025-tinymentalllms}. Researchers in psychology and computational linguistics emphasize the need for datasets that combine risk labels with structured analyses, such as user profiles or survey-based explanations. Such datasets better match real-world settings. Multifaceted insights can support clinical decision-making and downstream intervention.

Some recent efforts have introduced summarization-style datasets for depression. They leverage information extraction and automated summarization techniques~\cite{sotudeh-etal-2022-mentsum,sotudeh-etal-2021-tldr9,He2025Waste,shi2025medal}. However, these datasets often rely on model-generated content. The content is typically validated using automatic metrics such as ROUGE~\cite{lin-2004-rouge}. They also lack annotation by mental health professionals, which raises concerns about domain-specific reliability and accuracy.

To address these challenges, we introduce \textbf{CNSocialDepress} (\textbf{CNSD}), a publicly available Chinese-language dataset for depression risk detection that can be applied to the development of early-stage depression detection tools. It pairs binary risk labels with structured psychological analyses. All annotations are drafted and validated by certified mental health professionals, ensuring domain relevance and annotation quality. \textbf{CNSD} supports multiple task paradigms. These include binary classification, structured analysis generation, summarization, and fine-tuning large language models (LLMs) for psychological reasoning.

Our contributions are fourfold:
\begin{itemize}
    \item We release \textbf{CNSD}, a high-quality Chinese dataset for depression risk detection that combines binary labels with structured, interpretable psychological analyses.
    \item We propose an expert-in-the-loop annotation protocol with structured templates and quality control.
    \item We benchmark \textbf{CNSD} across diverse task settings, including classification, structured analysis generation, summarization, and LLM fine-tuning for psychological reasoning.
    \item We present and validate a pipeline for generating structured psychological analyses for depression risk (Section~\ref{sec:4}).
\end{itemize}

\section{Related Work}
\subsection{Existing Datasets for Depression Detection}

Current datasets for depression detection mainly come from English-language social media platforms such as Twitter, Reddit, Facebook, and Instagram~\cite{shen2017depression,erisk2022,zhang2021monitoring,raihan-etal-2024-mentalhelp,li2025chatmotion,jiang2025unihpr}. Common annotation strategies include identifying self-disclosure~\cite{yates-etal-2017-depression,bathina2021individuals,islam2018depression}, manual coding by clinical raters~\cite{1almouzini2019detecting,2yazdavar2020multimodal,3alhamed-etal-2024-classifying}, and symptom mapping~\cite{11seabrook2018predicting,22aldarwish2017predicting,33zhang-etal-2022-symptom,li2025wav2sem}. Symptom definitions are often based on DSM-5 criteria~\cite{edition2013diagnostic} and questionnaire instruments such as the PHQ-9~\cite{kroenke2001phq}.

Most datasets use binary labels. Others incorporate severity scales or symptom-specific tags~\cite{33zhang-etal-2022-symptom,mowery2017understanding,guan2025mcdi}. Multilingual resources also exist for depression detection. They cover languages such as Spanish~\cite{romero2024mentalriskes}, Arabic~\cite{maghraby2022modern}, Russian~\cite{stankevich2020depression}, Portuguese~\cite{santos2024setembrobr}, Japanese~\cite{yuka-niimi-2021-machine}, and Thai~\cite{hamalainen-etal-2021-detecting}. For Chinese, Sina Weibo is the most common data source~\cite{li2020automatic,shen2018cross,li2023mha,guo2023leveraging,yang2021fine,li2025aublendshape}. Datasets such as WU3D~\cite{wang-wu3d} and SWDD~\cite{cai2023depression} are widely used.

\subsection{Methodological Advancements}

Early work relied on statistical representations such as TF--IDF and hand-crafted features~\cite{yang2020big,li2022maskfpan}. These features were typically paired with traditional machine learning classifiers~\cite{cortes1995support,breiman2001random,mccallum1998comparison,dreiseitl2002logistic,he2026rareearth}. With the rise of neural approaches~\cite{Schmidhuber_2015}, representation learning became central to depression detection. Embedding methods such as Word2Vec~\cite{mikolov2013efficientestimationwordrepresentations} and neural architectures including LSTMs~\cite{sak2014long}, CNNs~\cite{kim-2014-convolutional}, Transformers~\cite{vaswani2017attention}, and BERT~\cite{devlin-etal-2019-bert} improved performance.

Recently, large language models (LLMs) have been explored for depression detection~\cite{lan2024depression,hu2024psycollmenhancingllmpsychological,MindChat,lai2023psy,li2026multiple,He2025Modular,shi2025explaining}. They can produce predictions and generate natural-language rationales or structured analyses, which supports interpretability. LLM-based approaches have also been used to improve explainability in mental health analysis~\cite{wang2024explainable,yang2024mentallama,xu2024mental,yang-etal-2023-towards,hu2024psycollmenhancingllmpsychological,gu2025mocount}.

\subsection{Summary and Gaps}

Despite these efforts, few datasets include structured psychological analyses, especially for non-English social media. Furthermore, available resources often lack annotation by mental health professionals, which limits practical utility. This study addresses these limitations by presenting an expert-validated Chinese-language dataset with detailed structured annotations. The dataset provides richer resources for depression risk detection and analysis.

\section{Construction of Dataset}
\label{sec:3}

\begin{figure*}[t]
    \centering
    \includegraphics[width=16cm]{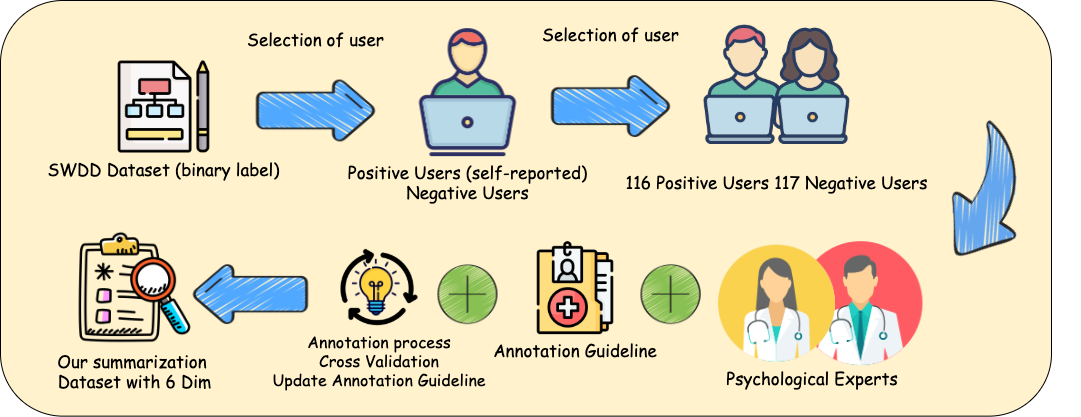}
    \caption{Dataset construction process. We sample 116 positive (depressed) users and 117 negative users from SWDD for expert annotation. Psychologists draft an initial guideline based on DSM-5 and corpus statistics, and iteratively refine it through random cross-validation. The final gold-standard data include a six-dimensional structured analysis summary for each user.}
    \label{fig:DCF}
\end{figure*}

Our raw data come from the SWDD dataset~\cite{cai2023depression}, a user-level corpus collected from Sina Weibo, one of the largest Chinese-language social media platforms. The dataset includes two user groups: depressed and non-depressed. Each user has dozens to hundreds of Weibo posts. In addition to binary user labels, SWDD provides expert-annotated user-level depression features based on DSM-5 criteria\footnote{\href{https://www.psychiatry.org/psychiatrists/practice/dsm/educational-resources/dsm-5-fact-sheets}{DSM-5 Fact Sheets.}}~\cite{edition2013diagnostic}.

From SWDD, we select 116 self-reported depressed (positive) users who disclosed a clinical diagnosis of depression in their posts, and 117 non-depressed (negative) users as candidates. Each selected user has at least 60 posts. The total length of a user’s posts is at least 3{,}000 tokens, computed using the Qwen2.5 tokenizer.\footnote{Qwen2.5 tokenizer.}

To guide annotation, a team of psychology experts drafted an initial guideline based on DSM-5 criteria, the PHQ-9~\cite{kroenke2001phq}, and prior work on linguistic markers of depression in online text~\cite{mothe:hal-03854902}.

We define two levels of criteria. Primary criteria cover objective statements with higher diagnostic specificity, such as clinical symptoms, medical records, and explicit self-reports. Secondary criteria capture subjective linguistic or emotional expressions with lower specificity, such as negative phrasing and emotional intensity. This hierarchy follows clinical principles that prioritize objective symptoms over subjective perceptions. The guidelines cover six dimensions:

\begin{itemize}
    \item \textbf{Dimension 1: Depressive psychological state (primary).}
    This dimension reflects loss of self-worth, overwhelming guilt, and suicidal ideation.
    Representative spans include \emph{inferiority}, \emph{apology}, \emph{death}, \emph{self-harm}, and \emph{suicide}.

    \item \textbf{Dimension 2: Medical expressions related to depression (primary).}
    This dimension covers mentions of medication use and clinical diagnoses, such as \emph{taking medication}, \emph{venlafaxine}, \emph{side effects}, \emph{depression}, \emph{anxiety disorder}, \emph{hospital}, and \emph{doctor}.

    \item \textbf{Dimension 3: Clinical symptoms related to depression (primary).}
    This dimension captures physiological or somatic changes, including appetite or weight changes, sleep disturbance, fatigue, and physical pain.
    Representative spans include \emph{appetite}, \emph{insomnia}, \emph{sleeping pills}, \emph{nightmares}, \emph{tiredness}, \emph{headache}, and \emph{stomachache}.

    \item \textbf{Dimension 4: Negative emotions (secondary).}
    This dimension covers adverse emotional states such as \emph{sadness}, \emph{grief}, \emph{anxiety}, \emph{loneliness}, and \emph{despair}.

    \item \textbf{Dimension 5: Potential external causes of depression (secondary).}
    This dimension captures possible triggers such as intimate relationship issues, family conflict, major social events, and everyday stressors.
    Representative spans include \emph{divorce}, \emph{parents}, \emph{school}, \emph{teacher}, and \emph{dropping out}.

    \item \textbf{Dimension 6: Language Use Patterns Related to Depression (secondary).}
    This dimension captures linguistic patterns such as negation, questions, and derogatory expressions.
    Representative spans include \emph{I don't know}, \emph{I don't like myself}, \emph{I don't want to}, \emph{I can't do it}, \emph{why}, and \emph{what should I do ?}.
\end{itemize}

\begin{figure}
    \centering
    \includegraphics[width=1\columnwidth]{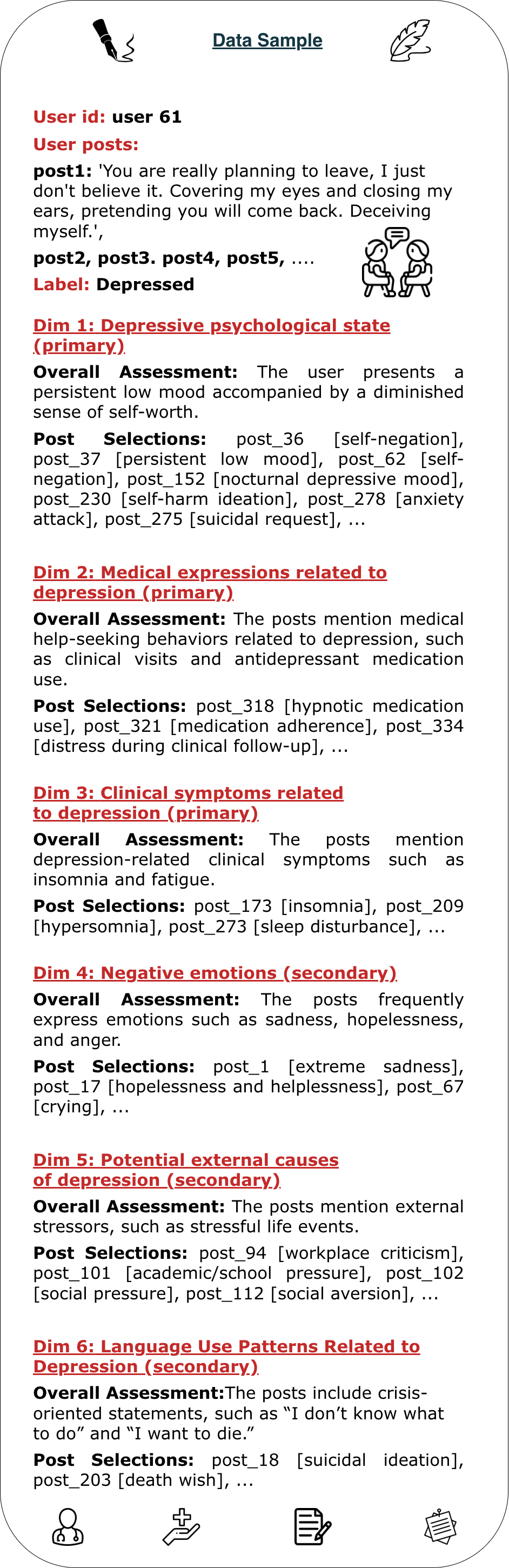}
    \caption{A user-level annotated entry from CNSD-Gold (English translations).}
    \label{fig:entry}
\end{figure}

Each user has multiple posts. We treat each post as the primary unit for review. Annotation is performed on semantic spans within each post. Each span is assigned to one of the six dimensions. A post may contain multiple spans, and different spans may belong to different dimensions.

Annotation was conducted by four senior researchers with expertise in psychology and depression scales. Annotators were blinded to the original binary labels in SWDD to ensure independent judgment. They periodically cross-checked overlapping samples to maintain consistency. Because the task relied more on qualitative judgment than on fixed quantitative criteria, the annotation guideline was not set in stone from the outset. It was revised in rounds as experts reviewed newly labeled data and resolved disagreements through discussion, and earlier annotations were corrected when the updated standards called for it. Given this evolving process, conventional inter-annotator agreement scores are not directly applicable. Since our annotation follows a different scheme from SWDD and all labeling was performed independently of the original labels, some user-level labels in our dataset differ from those in the original dataset.

This process yields the \textit{CNSD Gold} set (100 positive and 100 negative users) and the \textit{CNSD Test} set (16 positive and 17 negative users). Figure~\ref{fig:DCF} summarizes the construction process.

Each entry (Figure~\ref{fig:entry}) contains a binary label and a dimension-specific analysis. For each dimension, the analysis provides an overall assessment. It also lists post indices with brief justifications.

The overall statistics are presented in Table~\ref{tab:statistics}, and the 10{,}306 dimension-level annotations are summarized in Table~\ref{tab:dimension-stats}.

\begin{table}[t]
    \centering
    {\small
    \begin{tabular}{p{2cm} p{1.9cm} p{2.3cm}}
        \toprule
                       & \centering Positive (Depressive) & \centering Negative (Non-depressive) \tabularnewline
        \midrule
        NO. of Users  & \centering 116       & \centering 117       \tabularnewline
        NO. of Texts  & \centering 20,360    & \centering 23,818    \tabularnewline
        NO. of Tokens & \centering 1,024,978 & \centering 1,115,951 \tabularnewline
        \bottomrule
    \end{tabular}
    }
    \caption{Dataset statistics.}
    \label{tab:statistics}
\end{table}

\begin{table}[t]
    \centering
    {\small
    \begin{tabular}{c c c}
        \hline
        Dimension & Negative User & Positive User\\
        \hline
        Dim1 & 117   & 2127 \\
        Dim2 & 2     & 555  \\
        Dim3 & 126   & 768  \\
        Dim4 & 514   & 2933 \\
        Dim5 & 193   & 863  \\
        Dim6 & 231   & 1877 \\
        \hline
        Total & 1183  & 9123 \\
        \hline
    \end{tabular}
    }
    \caption{Dimension-level annotation statistics for negative and positive users. This table reports the number of expert-annotated depression-related spans in each dimension. A single text may contain multiple spans.}
    \label{tab:dimension-stats}
\end{table}

\section{Automated Dataset Generation Pipeline}
\label{sec:4}

Expert annotation requires substantial human effort and professional psychological expertise. As a result, user-level depression-risk datasets for Chinese social media remain scarce. To narrow this gap, we develop an automated dataset generation pipeline. We use it to construct \textit{CNSD Silver} from the SWDD binary classification dataset. The pipeline is informed by the manually annotated \textit{CNSD Gold} dataset (Section~\ref{sec:3}).

To build \textit{CNSD Silver}, we randomly sample 100 positive (depression-risk) users and 100 negative (non-risk) users from SWDD. Each user contributes dozens to hundreds of posts. We keep users with at least 3{,}000 tokens in total.

The pipeline consists of two modules.

\subsection{Module I: Dimension-Wise Automatic Labeling}

Module I uses a mid-sized model (Qwen2.5-14B~\cite{bai2023qwen}) to automatically assign posts to the six depression-related dimensions ($\mathcal{D} = \{D_1, D_2, \dots, D_6\}$). We fine-tune the model on text-level training data curated from positive users in \textit{CNSD Gold}. After fine-tuning, the model can assign each post to one or more dimensions.

\paragraph{Data preparation for fine-tuning Module I.}
\begin{enumerate}
    \item For each dimension $D_k$, we extract posts from positive users in \textit{CNSD Gold} that contain at least one annotated span of $D_k$. These posts form a set $S_k$.
    \item We identify the dimension with the fewest training posts. The minimum count is $n_{\text{min}} = 442$, which corresponds to \emph{depression-related medical expressions}:
    $$n_{\text{min}} = \min_{k \in \{1,2,\dots,6\}} n_k = 442.$$
    \item To balance dimensions, we downsample each $S_k$ to $n_{\text{min}}$ when $n_k > n_{\text{min}}$. This yields balanced sets $S'_k$:
    $$\small
    S'_k =
    \begin{cases}
    S_k, & \text{if } n_k = n_{\text{min}}, \\
    \text{RandomSubset}(S_k, n_{\text{min}}), & \text{if } n_k > n_{\text{min}}.
    \end{cases}
    $$
    We obtain $6 \times 442$ labeled positive posts. Each post is formatted using an instruction template such as: ``This post belongs to [\textit{category}] because it mentions [\textit{evidence}].'' We further augment the labels with 20 paraphrased expressions with the same meaning (Appendix~\ref{20_random_posi}).
    \item For negative training data, we split posts from negative users in \textit{CNSD Gold} into individual texts. We remove any text that contains depression-related content in any dimension. From the remaining texts, we uniformly sample $6 \times 442$ examples. We assign them negative label expressions indicating that there is no evidence for the target dimension (As with the positive examples, we create 20 negative label expressions to reduce overfitting; details are provided in~\ref{20_random_nega}).
    \item We fine-tune Qwen2.5-14B with LoRA~\cite{hu2022lora} on the combined set of $2 \times (6 \times 442)$ texts for 1 epoch. We use a learning rate of $5 \times 10^{-5}$, with the prompt provided in Appendix~\ref{instruction}.
\end{enumerate}

\subsection{Module II: Automatic Verification and Summarization}

In Module II, we apply the fine-tuned Module I model to label all posts for each user. We then collect posts that are assigned to at least one dimension. Next, we use DeepSeek-R1-671B~\cite{guo2025deepseek} to verify the assigned labels and the supporting evidence on a case-by-case basis. After verification, we summarize the collected posts using a few-shot prompting setup. The prompt instructs the model to act as a psychology expert and to produce structured summaries aligned with \textit{CNSD Gold}. We also provide detailed requirements and gold-standard examples. The full prompt is shown in Appendix~\ref{fig:module2}.

\section{Experiments}

We design our experiments to answer two research questions. \textbf{First}, how effectively does our data generation pipeline produce high-quality structured annotations ? \textbf{Second}, how well do various large language models (LLMs) perform on user-level depression risk summarization and structured psychological analysis ?

Based on these questions, we conduct three sets of experiments:
\begin{enumerate}
    \item To assess the quality of model-generated six-dimensional structured depression analyses. We compare models fine-tuned on \textit{CNSD Gold} and \textit{CNSD Silver} with baseline models and few-shot prompting.
    \item To evaluate the performance of these fine-tuned models on user-level depression risk classification.
    \item To benchmark leading LLMs on a structured depression analysis summarization task using \textit{CNSD Gold}.
\end{enumerate}

In Section~\ref{taskce}, we also use \textit{CNSD Gold} as a test set for binary depression classification. This further illustrates the dataset’s versatility.

\subsection{Baseline Models}

We use the following generative baselines: DeepSeek-R1-Distill-14B~\cite{guo2025deepseek}, Qwen2.5-14B~\cite{Qwen2.5}, GPT-4o\footnote{\url{https://openai.com/index/hello-GPT-4o/}}, GPT-4o-mini\footnote{\url{https://openai.com/index/GPT-4o-mini-advancing-cost-efficient-intelligence/}}, DeepSeek-R1-671B\footnote{\url{https://www.deepseek.com/}}~\cite{guo2025deepseek}, and Llama3-8B-Chinese-Chat~\cite{shenzhi_wang_2024}. We summarize key characteristics in Table~\ref{baseline_models}.

\subsection{Experimental Setup}

We use NVIDIA A800 (80GB) and A100 (80GB) GPUs. Unless otherwise specified, we set the generation temperature to 0.7. All fine-tuning experiments use the LLaMA-Factory framework~\cite{zheng2024llamafactory}. We apply LoRA~\cite{hu2022lora} for 1 epoch with a learning rate of $5 \times 10^{-5}$.

Our experiments have two parts. The first evaluates the pipeline for data generation. The second benchmarks models on the \textit{CNSD Gold} dataset for structured summarization and analysis generation.

For classification, we report accuracy, precision, recall, and F1-score. For generation, we report BLEU~\cite{papineni-etal-2002-bleu}, ROUGE-1~\cite{lin-2004-rouge}, and BERTScore~\cite{DBLP:conf/iclr/ZhangKWWA20}.

\subsection{Task I: Data Generation}
\label{taskI}

We conduct experiments on the \textit{CNSD Test} dataset. We focus on two Qwen2.5-14B models fine-tuned on \textit{CNSD Gold} and \textit{CNSD Silver}:
\begin{itemize}
    \item \textbf{Qwen2.5-14B Gold:} Fine-tuned on \textit{CNSD Gold}.
    \item \textbf{Qwen2.5-14B Silver:} Fine-tuned on \textit{CNSD Silver}. \textit{CNSD Silver} is generated using the pipeline in Section~\ref{sec:4} on 100 positive and 100 negative users sampled from SWDD.
\end{itemize}

We compare these models with other baselines in two aspects:
\begin{enumerate}
    \item Depression risk classification performance.
    \item Quality of the generated six-dimensional analyses, including hallucinated or unsupported content.
\end{enumerate}

Unless noted otherwise, we use user-level generation. We concatenate all posts from a user as input. We instruct the model to output a binary label and a corresponding justification.

\begin{table*}[t]
\centering
\small
\begin{tabular}{lccc}
\toprule

\textbf{Strategy} & \textbf{BERTScore} & \textbf{ROUGE-1} & \textbf{BLEU} \\
\midrule
\textbf{Pipeline} & \textbf{0.791} & \textbf{0.478} & \textbf{0.288} \\
FS:Qwen2.5 14B & 0.649 & 0.076 & 0.075 \\
FS:GPT-4o & 0.678 & 0.269 & 0.094 \\
FS:GPT-4o Mini & 0.674 & 0.170 & 0.070 \\
FS:DeepSeek R1 671B & 0.678 & 0.301 & 0.054 \\
FS:DeepSeek R1 Distill -14B & 0.6756 & 0.191 & 0.073 \\
\bottomrule
\end{tabular}%
\caption{Comparison of generation strategies on text quality. In the table, FS stands for Few-Shot. Pipeline refers to the automated dataset generation pipeline proposed in Section~\ref{sec:4}, applied to our proposed dataset in Section~\ref{sec:3}. The purpose is to show that, with the same prompt, our pipeline achieves the highest text generation quality while meeting our task requirements. The prompt used in this experiment is shown in Figure~\ref{fig:prompt_table}.}
\label{p1_dg}
\end{table*}

\begin{figure}[t]
    \centering
    \includegraphics[width=0.9\linewidth]{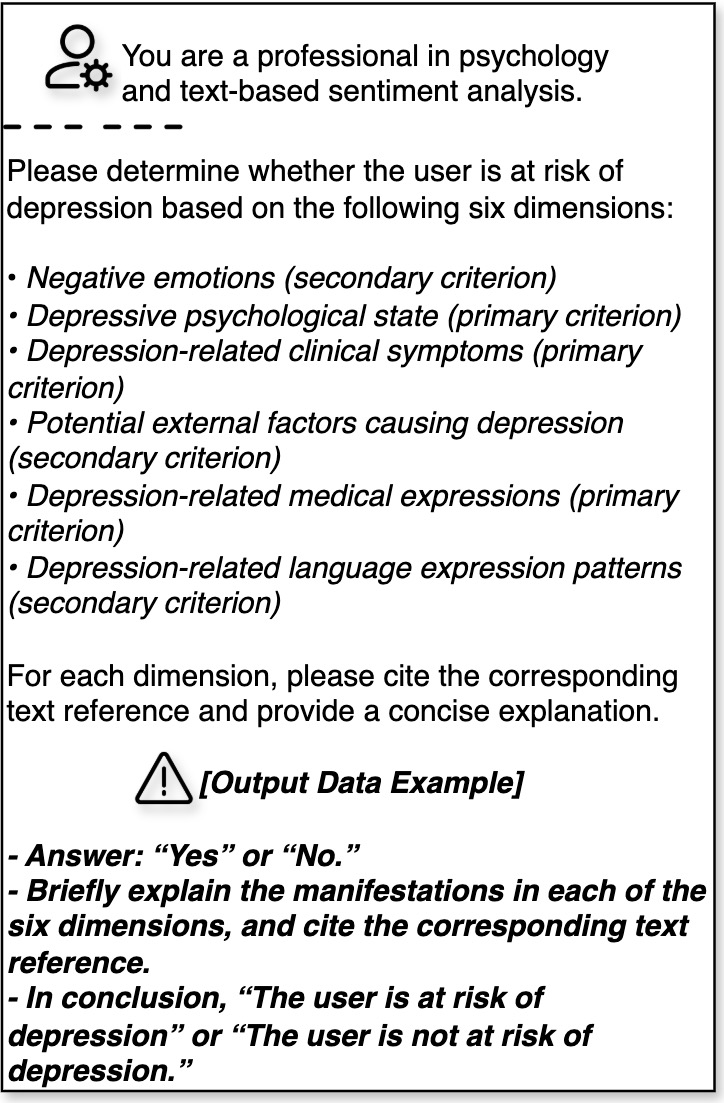}
    \caption{Prompt used for Table \ref{p1_dg}.}
    \label{fig:prompt_table}
\end{figure}

\subsubsection{Six-Dimensional Structured Depression Analysis}
\label{sec5.3.1}

Prior work suggests that incorporating structured knowledge can improve LLM outputs~\cite{moiseev-etal-2022-skill}. \textit{CNSD Gold} and \textit{CNSD Silver} provide fine-grained structured annotations across six dimensions. They support user-level summarization and analysis of social media content. We hypothesize that fine-tuning on these datasets can reduce hallucinations. It may also improve generation quality compared to few-shot prompting and larger models.

Table~\ref{p1_dg} compares different generation strategies using automatic metrics. Our Pipeline achieves the best scores on all three metrics: BERTScore (0.791), ROUGE-1 (0.478), and BLEU (0.288). These results indicate stronger semantic alignment, higher lexical overlap, and better $n$-gram precision. Few-shot (FS) prompting yields lower scores across models, including Qwen2.5-14B, GPT-4o, and DeepSeek-R1-671B. Notably, DeepSeek-R1-671B does not outperform the Pipeline. This suggests that the proposed Pipeline produces text that matches the reference more closely than few-shot prompting, across both smaller and larger models.

We also compare the original Qwen2.5-14B model with Qwen2.5-14B Gold and Qwen2.5-14B Silver (Table~\ref{tab:11}). We use two complementary evaluation methods.

\begin{itemize}
    \item \textbf{Automatic metrics.} The original model achieves a BERTScore of 0.649. It increases to 0.764 for Gold (+16.9\%) and 0.787 for Silver (+21.5\%). ROUGE-1 rises from 0.076 to 0.217 for Gold (+172\%) and 0.218 for Silver (+172\%). BLEU improves from 0.075 to 0.186 for Gold (+149\%) and 0.237 for Silver (+218\%).
    \item \textbf{Human evaluation.} Two linguistic experts evaluate accuracy, coverage, and hallucination. Compared to the original Qwen2.5-14B, both Gold and Silver improve substantially. Silver increases by 33.4\% in accuracy and 37.1\% in coverage. Its hallucination score increases by 15.5\% (higher is better). Gold achieves larger gains. Accuracy increases by 42.9\% and coverage increases by 58.0\%. The hallucination score increases by 26.8\%. Overall, Gold performs best in human evaluation. Silver is slightly behind but remains close to Gold and clearly outperforms the original model.
\end{itemize}

In summary, Gold performs best in human evaluation. Silver performs better on automatic metrics. Both models substantially outperform the original model. These results show that fine-tuning with our dataset and pipeline can improve generation quality at semantic, lexical, and structural levels.

\begin{table*}[t]
    \centering
    \small
    \begin{tabular}{lccccccc}
        \toprule
        Model & BERTScore & ROUGE-1 & BLEU & Human.acc & Human.cov & Human.hallu \\
        \midrule
        Qwen2.5-14B         & 0.649 & 0.076 & 0.075 & 0.583 & 0.574 & 0.657 \\
        Qwen2.5-14B Silver  & \textbf{0.787} & \textbf{0.218} & \textbf{0.237} & 0.778 & 0.787 & 0.759 \\
        Qwen2.5-14B Gold    & 0.764 & 0.217 & 0.186 & \textbf{0.833} & \textbf{0.907} & \textbf{0.833} \\
        \bottomrule
    \end{tabular}
    \caption{Comparison of generation quality across models. \texttt{Qwen2.5-14B Gold} is fine-tuned on \textit{CNSD Gold}. \texttt{Qwen2.5-14B Silver} is fine-tuned on \textit{CNSD Silver}, which is generated by our pipeline (Section~\ref{sec:4}). We report automatic metrics (BERTScore, ROUGE-1, BLEU) and human evaluation metrics: \textit{acc} (accuracy), \textit{cov} (content coverage), and \textit{hallu} (hallucination; higher is better). Higher values indicate better performance for all metrics.}
    \label{tab:11}
\end{table*}

\subsubsection{Classification Task}

In the depression classification task, \textit{Qwen2.5-14B Silver} achieves the best performance among all models. It reaches an accuracy of 0.944 and an F1 score of 0.941. It also surpasses \textit{Qwen2.5-14B Gold}, which is fine-tuned on human-annotated data. Moreover, it outperforms large-scale models such as GPT-4o (accuracy = 0.917, F1 = 0.923) and DeepSeek-R1-671B (accuracy = 0.861, F1 = 0.872). These results support the effectiveness of our generation pipeline and suggest that the resulting silver data are of high quality. Detailed results are reported in Table~\ref{tab:1}.

\begin{table*}[t]
\centering
\small
\setlength{\tabcolsep}{1.5pt}
\begin{tabular}{lcccc}
\toprule
\textbf{Model} & \textbf{Accuracy} & \textbf{Recall} & \textbf{Precision} & \textbf{F1} \\
\midrule
Qwen2.5-14B & 0.889 & 0.944 & 0.850 & 0.894 \\
Qwen2.5-14B (FT, Silver) & \textbf{0.944} & 0.889 & \textbf{1.000} & \textbf{0.941} \\
Qwen2.5-14B (FT, Gold) & 0.861 & 0.889 & 0.842 & 0.864 \\
GPT-4o & 0.917 & \textbf{1.000} & 0.857 & 0.923 \\
GPT-4o-mini & 0.667 & \textbf{1.000} & 0.600 & 0.750 \\
DeepSeek-R1-671B & 0.861 & 0.944 & 0.801 & 0.872 \\
DeepSeek-R1-Distill-14B & 0.889 & \textbf{1.000} & 0.818 & 0.900 \\
\bottomrule
\end{tabular}
\caption{Comparison of Models for Generated Dataset Quality: Application to Classification Tasks.}
\label{tab:1}
\end{table*}

\subsection{Task II: Structured Summarization and Analysis Generation}
\label{task2}

Our dataset supports research on generative models for depression risk assessment from social media. It is particularly useful for user-level summarization of depressive signals. To the best of our knowledge, there is no directly comparable open-source dataset with the same annotation schema. We therefore conduct an initial benchmark of mainstream generative models on our annotated dataset of 233 users. Results are reported in Table~\ref{tab:generation_test}. 

We use the following prompt:

\begin{itemize}
    \item \noindent\textbf{Instruction:}
    \textit{Read the following texts written by one user. Then answer the question.}

    \item \noindent\textbf{Question:} Based on these texts, does this user exhibit a depressive mood ?
    
    \begin{itemize}
    \item If the user exhibits a depressive mood, answer \texttt{Yes}.
    \item If the user does not exhibit a depressive mood, answer \texttt{No}.
    \end{itemize}

    \item \noindent\textbf{Output requirements:} Provide a brief justification based on evidence from the texts. Then produce a structured summary aligned with the annotation schema.
\end{itemize}

The results vary across models. BERTScore falls in a narrow range (0.654--0.710), with GPT-4o achieving the highest score. ROUGE-1 is highest for DeepSeek-R1-671B (0.454), which exceeds the second-best model (GPT-4o at 0.346) by an absolute margin of 0.108. BLEU ranges from 0.093 (Qwen2.5-7B) to 0.152 (GPT-4o).

\begin{table}[htbp]
\centering
\small
\begin{tabularx}{\linewidth}{@{}l *{3}{>{\centering\arraybackslash}X}@{}}
\toprule
Model & BERT Score & ROUGE-1 & BLEU \\
\midrule
Llama3-8B & 0.654 & 0.194 & 0.122 \\
GLM4-9B-Chat & 0.694 & 0.208 & 0.140 \\
Qwen2.5-7B & 0.680 & 0.208 & 0.093 \\
DeepSeek-R1-671B & 0.698 & \textbf{0.454} & 0.125 \\
GPT-4o-mini & 0.696 & 0.166 & 0.126 \\
GPT-4o & \textbf{0.710} & 0.346 & \textbf{0.152} \\
\bottomrule
\end{tabularx}
\caption{Model performance on the depression risk summarization and analysis generation task.}
\label{tab:generation_test}
\end{table}

\subsection{Task III: Classification Experiments}
\label{taskce}

Our dataset is primarily designed for generation tasks. We also conduct a binary classification experiment for depression risk. In addition to our dataset, we use the original SWDD and WU3D datasets. For each dataset, we randomly sample 200 positive and 200 negative users.

Besides the generative models described in Appendix~\ref{baseline_models}, we evaluate BERT-based classifiers~\citep{devlin-etal-2019-bert}. Implementation details are provided in Appendix~\ref{bert_model}. Results are reported in Appendix~\ref{section_tab_class_e}.

\section{Conclusion}

We introduce \textbf{CNSocialDepress}, a new dataset for depression risk analysis from Chinese social media. To our knowledge, it is among the first to provide both user-level and text-level labels. The dataset combines binary depression indicators with fine-grained six-dimensional analyses annotated by mental health professionals. We also develop an automated data generation pipeline. It reduces manual annotation effort and supports language model fine-tuning for classification and summarization. We expect \textbf{CNSocialDepress} to be useful for a range of research and applications, supporting early risk identification and downstream mental health interventions.

\section{Limitations}

While \textbf{CNSocialDepress} provides a valuable resource for depression risk analysis on Chinese social media, several limitations should be noted. First, the dataset is sourced exclusively from Weibo. This may introduce platform-specific demographic biases, such as the underrepresentation of rural or elderly populations.

Second, linguistic diversity is not fully captured. This includes dialectal variation and metaphorical expressions that are common in Chinese depressive discourse. This limitation may reduce generalizability across regions. Third, expert annotation ensures high quality, but it limits scalability. It also results in a relatively small dataset.

In addition, the current structured analyses focus primarily on depressive symptoms. They omit comorbid mental health conditions (e.g., anxiety) and broader social context. Future work will expand data collection to multiple platforms. We also plan to update lexicons to capture emerging expressions. Finally, we will explore hybrid annotation frameworks (e.g., expert-guided crowdsourcing) to improve coverage and efficiency. Privacy-preserving techniques and longitudinal tracking can further strengthen ethical and practical utility.

\section{Ethics Statement}

This dataset is built upon the SWDD benchmark~\cite{cai2023depression}, released in early 2023. The original data consist of user-level posts from Weibo. They were annotated by psychology experts based on DSM-5 criteria.

During re-annotation, we applied a second round of de-identification to protect privacy. We remove personal information, geographic locations, and other potential identifiers from the texts. We follow recommended data protection practices~\citet{benton2017ethical} and comply with the GDPR (General Data Protection Regulation).\footnote{\url{https://eur-lex.europa.eu/legal-content/EN/TXT/PDF/?uri=CELEX:02016R0679-20160504}} We also recruit experienced researchers in depression-related psychology to provide expert annotations.

This dataset is intended for research on depression risk identification and for building assistive models. It can support multi-dimensional analysis of risk-related signals and the generation of structured summaries. It is not designed for clinical diagnosis, triage, or treatment decisions. Any outputs should be used only as references for professionals or as a self-assessment aid.

No algorithmic system can replace in-person psychiatric evaluation or provide a definitive diagnosis. If a system built on this dataset flags a high-risk case, users should be encouraged to seek professional help. The same applies to individuals with persistent mood disturbances or impaired functioning. Online self-assessment tools cannot substitute for clinical evaluation.

\section{Bibliographical References}
\nocite{*}
\bibliographystyle{lrec2026-natbib}

\bibliography{lrec2026-example}

\newpage
\appendix
\section{Appendix}
\label{sec:appendix}

\subsection{Baseline Models}
\label{baseline_models}
\begin{itemize}
    \item \textbf{DeepSeek-R1} is a large-scale model designed for complex reasoning. It performs well on tasks such as mathematics and programming, while remaining competitive on general-purpose benchmarks. We include it as a strong reasoning-oriented baseline for structured psychological analysis and depression risk assessment.

    \item \textbf{DeepSeek-R1-Distill-14B} is a distilled reasoning model derived from Qwen2.5-14B via knowledge distillation. It aims to retain key reasoning capabilities at lower computational cost. We include it as an efficient baseline for depression risk analysis.

    \item \textbf{Qwen2.5-14B} is a large language model from the Qwen series. It provides strong instruction following and long-form generation. We include it as a Chinese-capable baseline for generating structured psychological analyses from social media posts.

    \item \textbf{GPT-4o} is a multimodal large language model released by OpenAI. It supports multilingual generation, including Chinese. We include it as a strong general-purpose baseline for depression detection and psychological analysis.

    \item \textbf{GPT-4o-mini} is a smaller model in the GPT-4o family. It is designed for efficiency and lower inference cost. We include it as a lightweight baseline for depression risk identification and structured summarization.

    \item \textbf{Llama3-8B-Chinese-Chat}, derived from Meta-Llama-3-8B, is an instruction-tuned model for Chinese dialogue. It supports both Chinese and English inputs. We include it as a Chinese dialogue-oriented baseline for user-level depression risk analysis and generation.
\end{itemize}

\begin{CJK}{UTF8}{gbsn}
\subsection{20 Randomly Generated Expression for the Positive Users}
\label{20_random_posi}
\begin{enumerate}
    \item Carefully read the complete user Weibo text below and extract key information across six specified dimensions according to the content. Also pay attention to the user's depressive state, ensuring that each dimension's explanation is consistent with this state. \\
    \textbf{[Input]} User depressive state: \{label\}, Full Weibo text: \{text\}. \\
    \textbf{[Output requirements]} - For each dimension, if evidence exists in the text, provide the text index (e.g., text\_xx, 文本\_xx) and a brief note (e.g., text\_xx[sad]). - If no evidence is found for a dimension, state clearly that no relevant evidence was found. \\
    Note: Only output text index and explanation, not the actual text.
    
    \item Read the entire user Weibo text below and, based on its content, extract core information under the following six dimensions. Also consider the user's depressive status, ensuring that each dimension's description matches this state. \\
    \textbf{[Input]} User depressive state: \{label\}, Full Weibo text: \{text\}. \\
    \textbf{[Output requirements]} - For each dimension, if relevant information is detected, provide the text index (e.g., text\_xx, 文本\_xx) and a short note (e.g., text\_xx[unhappy]). - For dimensions without evidence, explicitly state no relevant evidence found. \\
    Note: Output should include only text index and explanation, not the actual text.
    
    \item Read through the complete user Weibo text below and extract key information across six dimensions based on its content. Pay close attention to the user's depressive state, ensuring that the explanations are consistent with it. \\
    \textbf{[Input]} User depressive state: \{label\}, Full Weibo text: \{text\}. \\
    \textbf{[Output requirements]} - If relevant evidence is found in a dimension, provide the text index (e.g., text\_xx, 文本\_xx) and a short note (e.g., text\_xx[sorrow]). - If no evidence is found, indicate explicitly. \\
    Note: Provide only indices and explanations, not the original content.
    
    \item Carefully read the full Weibo text below and extract key evidence across six dimensions according to the text. At the same time, take into account the user's depressive state, ensuring that explanations for each dimension align with this state. \\
    \textbf{[Input]} User depressive state: \{label\}, Full Weibo text: \{text\}. \\
    \textbf{[Output requirements]} - For each dimension, if there is corresponding evidence, output the text index (e.g., text\_xx, 文本\_xx) and a related note (e.g., text\_xx[frustrated]). - If no evidence is found, state that no evidence exists. \\
    Note: Only provide text indices and notes, not the raw text.
    
    \item Please read the full Weibo text below and extract key information from six dimensions. At the same time, pay attention to the user's depressive state, ensuring that your explanations match this state. \\
    \textbf{[Input]} User depressive state: \{label\}, Full Weibo text: \{text\}. \\
    \textbf{[Output requirements]} - For each dimension, if evidence is detected, output the text index (e.g., text\_xx, 文本\_xx) and a short description (e.g., text\_xx[sadness]). - If no evidence is found, specify that no evidence was found. \\
    Note: Only output text indices and explanations, not the original text.
    
    \item Read the complete Weibo text below carefully and extract key information across six dimensions based on the text. Pay attention to the user's depressive state and ensure that your descriptions are consistent with it. \\
    \textbf{[Input]} User depressive state: \{label\}, Full Weibo text: \{text\}. \\
    \textbf{[Output requirements]} - If relevant evidence is found, provide the text index (e.g., text\_xx, 文本\_xx) and a brief description (e.g., text\_xx[low mood]). - If no evidence is found, clearly state that no evidence exists. \\
    Note: Only provide text indices and explanations, not the original text.
    
    \item Please read through the complete user Weibo text below and extract key information across six dimensions based on its content. Pay attention to the user's depressive state, ensuring that each explanation is aligned with this state. \\
    \textbf{[Input]} User depressive state: \{label\}, Full Weibo text: \{text\}. \\
    \textbf{[Output requirements]} - For each dimension, if evidence exists, provide the text index (e.g., text\_xx, 文本\_xx) and a short note (e.g., text\_xx[loss]). - If no evidence is found, clearly state no relevant evidence. \\
    Note: Only provide text indices and notes, not the original text.
    
    \item Read the Weibo text below and extract key evidence across six dimensions according to the content, while considering the user's depressive state to ensure explanations are consistent. \\
    \textbf{[Input]} User depressive state: \{label\}, Full Weibo text: \{text\}. \\
    \textbf{[Output requirements]} - If a dimension contains relevant evidence, provide the text index (e.g., text\_xx, 文本\_xx) and a short note (e.g., text\_xx[low mood]). - If no evidence is found, state explicitly that no evidence exists. \\
    Note: Output only text indices and explanations, not the original text.
    
    \item Read the complete Weibo text carefully and extract key evidence across six dimensions based on the text. Pay attention to the user's depressive state, ensuring that explanations match this state. \\
    \textbf{[Input]} User depressive state: \{label\}, Full Weibo text: \{text\}. \\
    \textbf{[Output requirements]} - For each dimension, if relevant evidence is found, output the text index (e.g., text\_xx, 文本\_xx) and a brief explanation (e.g., text\_xx[depressed]). - If no evidence is found, state explicitly that there is none. \\
    Note: Provide only text indices and notes, not the original text.
    
    \item Read through the entire Weibo text and extract core information across six dimensions according to the content. Pay attention to the user's depressive state and ensure that the descriptions are consistent with this state. \\
    \textbf{[Input]} User depressive state: \{label\}, Full Weibo text: \{text\}. \\
    \textbf{[Output requirements]} - For each dimension, if relevant evidence exists, output the text index (e.g., text\_xx, 文本\_xx) and a short explanation (e.g., text\_xx[pessimistic]). - If no evidence is found, state clearly that no evidence exists. \\
    Note: Only output text indices and explanations, not the original content.
    
    \item Please read the user Weibo text below completely and extract essential information across six dimensions. Ensure that the explanations for each dimension align with the depressive state given. \\
    \textbf{[Input]} User depressive state: \{label\}, Full Weibo text: \{text\}. \\
    \textbf{[Output requirements]} - If relevant evidence is detected, list the text index (e.g., text\_xx, 文本\_xx) and a short description (e.g., text\_xx[melancholy]). - If no evidence is found, specify that no evidence was detected. \\
    Note: Output should only contain indices and notes, not the actual text.
    
    \item Read carefully the full Weibo text provided and extract key evidence from six dimensions. Pay attention to the user's depressive state, ensuring that outputs remain consistent with this state. \\
    \textbf{[Input]} User depressive state: \{label\}, Full Weibo text: \{text\}. \\
    \textbf{[Output requirements]} - If a dimension contains evidence, output the text index (e.g., text\_xx, 文本\_xx) and a concise note (e.g., text\_xx[disheartened]). - If no evidence exists, specify so. \\
    Note: Output indices and explanations only, not original content.
    
    \item Please go through the complete Weibo text below and identify critical information across six dimensions. Ensure the notes correspond with the user's depressive condition. \\
    \textbf{[Input]} User depressive state: \{label\}, Full Weibo text: \{text\}. \\
    \textbf{[Output requirements]} - Where evidence is found, list the text index (e.g., text\_xx, 文本\_xx) with a short explanation (e.g., text\_xx[downhearted]). - If no evidence is detected, indicate explicitly. \\
    Note: Provide only text indices and notes, excluding original text.
    
    \item Read carefully the following Weibo text and extract evidence across six preset dimensions. Make sure the explanation for each aligns with the depressive state. \\
    \textbf{[Input]} User depressive state: \{label\}, Full Weibo text: \{text\}. \\
    \textbf{[Output requirements]} - If evidence exists for a dimension, output the text index (e.g., text\_xx, 文本\_xx) and a short remark (e.g., text\_xx[low mood]). - If no evidence is found, specify clearly. \\
    Note: Output only indices and remarks, not the raw text.
    
    \item Read through the following Weibo text and extract major evidence across six given dimensions. Ensure the output is consistent with the depressive state specified. \\
    \textbf{[Input]} User depressive state: \{label\}, Full Weibo text: \{text\}. \\
    \textbf{[Output requirements]} - If evidence is found, provide the text index (e.g., text\_xx, 文本\_xx) with a brief note (e.g., text\_xx[hopeless]). - If no evidence appears, state clearly. \\
    Note: Only provide indices and notes, not original text.
    
    \item Please carefully review the Weibo text below and extract key information from six preset dimensions. All notes must align with the depressive condition of the user. \\
    \textbf{[Input]} User depressive state: \{label\}, Full Weibo text: \{text\}. \\
    \textbf{[Output requirements]} - If there is evidence, output the corresponding text index (e.g., text\_xx, 文本\_xx) and a short remark (e.g., text\_xx[emotional slump]). - If not found, specify clearly that none exists. \\
    Note: Only include indices and notes, no original content.
    
    \item Please go through the complete Weibo text and extract the critical information across six given dimensions. Make sure your notes are consistent with the depressive state provided. \\
    \textbf{[Input]} User depressive state: \{label\}, Full Weibo text: \{text\}. \\
    \textbf{[Output requirements]} - For each dimension, if evidence is found, provide the text index (e.g., text\_xx, 文本\_xx) with a concise explanation (e.g., text\_xx[feeling low]). - If no evidence exists, indicate so. \\
    Note: Only output indices and explanations, not actual text.
    
    \item Carefully read the Weibo text provided and extract important evidence across six specified dimensions. Keep the explanations aligned with the depressive state. \\
    \textbf{[Input]} User depressive state: \{label\}, Full Weibo text: \{text\}. \\
    \textbf{[Output requirements]} - If a dimension contains evidence, state the text index (e.g., text\_xx, 文本\_xx) and a short remark (e.g., text\_xx[upset]). - If not, state that no evidence is found. \\
    Note: Output only indices and remarks, not text.
    
    \item Read the Weibo text completely and extract relevant evidence from six dimensions. Ensure each dimension's note is consistent with the depressive state given. \\
    \textbf{[Input]} User depressive state: \{label\}, Full Weibo text: \{text\}. \\
    \textbf{[Output requirements]} - If relevant evidence is detected, provide the text index (e.g., text\_xx, 文本\_xx) and short description (e.g., text\_xx[distressed]). - If not found, state clearly no evidence. \\
    Note: Only indices and notes, exclude text itself.
    
    \item Please read through the following full Weibo text and extract essential evidence across six preset dimensions. Ensure all notes are consistent with the depressive state. \\
    \textbf{[Input]} User depressive state: \{label\}, Full Weibo text: \{text\}. \\
    \textbf{[Output requirements]} - If evidence is found, provide the text index (e.g., text\_xx, 文本\_xx) and a short explanation (e.g., text\_xx[worry]). - If no evidence is detected, state that explicitly. \\
    Note: Only indices and explanations, not original text.
\end{enumerate}
\end{CJK}

\subsection{20 Distinct Label Expressions for the Negative Users}
\label{20_random_nega}
\begin{enumerate}
    \item "This text does not contain any expressions of negative emotions, nor does it involve any dimensions of depression."
    \item "There are no descriptions related to negative emotions in the text, so it does not fit any depression-related dimensions."
    \item "This text does not include expressions of negative emotions, and therefore it is not classified under any depression dimensions."
    \item "The content does not reflect negative emotions and does not belong to any depression indicators."
    \item "This passage contains no expressions of negative emotions and is not applicable to any depression dimensions."
    \item "There are no descriptions of negative emotions in the text, so it is not classified under any depression-related dimensions."
    \item "This content does not exhibit negative emotions and therefore does not involve any depression dimensions."
    \item "The text does not contain any expressions related to negative emotions, nor does it meet the criteria for depression dimensions."
    \item "There are no signs of negative emotions in this text, so it does not belong to any depression dimension."
    \item "The text does not describe negative emotions and does not cover any depression-related dimensions."
    \item "This passage does not contain any expressions related to negative emotions and is not within the scope of depression dimensions."
    \item "The content does not include expressions of negative emotions, so it does not fall into any depression dimensions."
    \item "There are no expressions of negative emotions in the text, nor does it meet the criteria for depression dimensions."
    \item "This text does not exhibit any negative emotions and does not involve any depression dimensions."
    \item "This content does not include descriptions of negative emotions and therefore is not classified under any dimensions of depression."
    \item "The text does not show any negative emotions and is not applicable to depression-related dimensions."
    \item "No expressions of negative emotions are seen in this passage, so it does not meet any depression dimension standards."
    \item "There are no expressions of negative emotions in the text, so it does not belong to any depression dimension."
    \item "This text does not display any expressions related to negative emotions and does not cover the dimensions of depression."
    \item "The content lacks descriptions of negative emotions and is therefore not classified under any depression dimensions."
\end{enumerate}

\subsection{Instructions for the Fine-Tuning Process}
\label{instruction}
\begin{itemize}
    \item \textbf{Complete Instruction = Task Instruction + Supplementary Instruction}
\end{itemize}

\textbf{Task Instruction:}

This task involves the following 6 dimensions of depression:
\begin{itemize}
    \item Potential External Causes of Depression (Secondary Judgment Criterion)
    \item Depression-Related Clinical Symptoms (Primary Judgment Criterion)
    \item Depression-Related Language Expression Patterns (Secondary Judgment Criterion)
    \item Depression-Related Medical Expressions (Primary Judgment Criterion)
    \item Depressive Psychological State (Primary Judgment Criterion)
    \item Negative Emotions (Secondary Judgment Criterion)
\end{itemize}

\textbf{Supplementary Instructions (randomly select one):}

\begin{enumerate}
    \item "Please read the following text segment and determine whether it contains any of the above 6 expressions of depressive emotion, and provide a brief explanation; if none is present, please indicate that the text does not belong to any depression dimension."
    \item "Read the following text and analyze whether it demonstrates any one of the aforementioned 6 depressive emotion expression dimensions, and provide a brief explanation; if such expression is absent, please indicate that the text does not correspond to any dimension."
    \item "Please carefully read the following text segment and confirm whether any one of the 6 depressive emotion expression dimensions mentioned above exists, and include a brief explanation; if not, please state that the text does not belong to any dimension."
    \item "Please read the following content and determine whether it embodies any one of the above 6 depressive emotion expression dimensions, and provide a brief explanation; if not, please indicate that the text does not cover any depression dimension."
    \item "Read the following text segment and determine whether it contains any one of the 6 depressive emotion expressions mentioned above, and provide a concise explanation; if not, please state that the text does not belong to any depression dimension."
    \item "Please read the following text and determine whether any one of the above 6 depressive emotion expression dimensions appears, and include a brief explanation; if not, please indicate that the text does not meet any dimension."
    \item "Read the following text and check whether any one of the above 6 depressive emotion expression dimensions is presented, and provide a brief explanation; if not, please state that the text does not belong to any dimension."
    \item "Please review the following text segment and determine whether any one of the above 6 depressive emotion expression dimensions exists, and provide a brief explanation; if not, then indicate that the text does not belong to any dimension."
    \item "Read the following text and confirm whether any one of the above 6 depressive emotion expression dimensions is embodied, and provide a brief explanation; if not, please state that the text does not involve any depression dimension."
    \item "Please read the following content and determine whether it contains any one of the 6 depressive emotion expression dimensions mentioned earlier, and provide a brief explanation; if not, please indicate that the text does not belong to any dimension."
    \item "Read the following text segment and verify whether it demonstrates any one of the above 6 depressive emotion expression dimensions, and include a brief explanation; if not, please state that the text does not meet any dimension."
    \item "Please read the following text and check whether it exhibits any one of the above 6 depressive emotion expression dimensions, and provide a brief explanation; if not, please indicate that the text does not cover any depression dimension."
    \item "Read the following text and confirm whether any one of the 6 depressive emotion expressions mentioned above exists, and include a brief explanation; if such a feature is absent, please state that the text does not belong to any dimension."
    \item "Please carefully read the following text segment and determine whether it embodies any one of the above 6 depressive emotion expression dimensions, and provide a concise explanation; if not, please indicate that the text does not correspond to any dimension."
    \item "Read the following text segment and confirm whether it contains any one of the above 6 depressive emotion expression dimensions, and provide a brief explanation; if not, please state that the text does not involve any depression dimension."
    \item "Please read the following text segment and determine whether any one of the 6 depressive emotion expression dimensions mentioned above appears in the text, and include a brief explanation; if not, please indicate that the text does not belong to any dimension."
    \item "Read the following content and check whether it contains any one of the above 6 depressive emotion expression dimensions, and provide a brief explanation; if not, please state that the text does not belong to any depression dimension."
    \item "Please read the following content and determine whether any one of the above 6 depressive emotion expression dimensions exists, and provide a brief explanation; if not, please indicate that the text does not involve any dimension."
    \item "Read the following text segment and confirm whether it demonstrates any one of the above 6 depressive emotion expression dimensions mentioned above, and provide a brief explanation; if not, please state that the text does not meet any dimension."
    \item "Please read the following text and determine whether it contains any one of the above 6 depressive emotion expression dimensions, and provide a concise explanation; if not, please indicate that the text does not belong to any depression dimension."
\end{enumerate}

\subsection{BERT-based Models}
\label{bert_model}

\textbf{BERT-base-chinese}\footnote{\url{https://huggingface.co/google-bert/bert-base-chinese}} follows the BERT-Base architecture~\cite{devlin-etal-2019-bert} and is pre-trained on large-scale Chinese corpora. It uses masked language modeling (MLM) and next sentence prediction (NSP). The model learns bidirectional representations for Chinese text, typically at the character level.

\textbf{Chinese-RoBERTa-wwm-ext}~\cite{chinese-BERT-wwm} is based on RoBERTa~\cite{liu2019roberta} and uses whole word masking (WWM). It masks complete words rather than individual characters during MLM. This design can better capture word-level semantics in Chinese. The extended pre-training (``ext'') on large Chinese corpora further improves robustness across NLP tasks.

\textbf{StructBERT-mental}~\cite{wang2019structbert} builds on StructBERT, which incorporates structural objectives during pre-training. It emphasizes word- and sentence-level structure. In addition, it is adapted for mental health text, which can benefit mental health-related classification.

\begin{table*}[b]
\subsection{Classification Performance Comparison}
\label{section_tab_class_e}
\centering
\small
\begin{tabular}{lcc|cc|cc}
\toprule
{\textbf{Model/Test Dataset}} & \multicolumn{2}{c|}{\textbf{Swdd Origin}} & \multicolumn{2}{c|}{\textbf{Wu3d}} & \multicolumn{2}{c}{\textbf{CNSD (Gold)}} \\
& \textbf{Acc} & \textbf{F1} & \textbf{Acc} & \textbf{F1} & \textbf{Acc} & \textbf{F1} \\
\midrule
RoBERTa Chinese (trained by swdd origin) & 0.89 & 0.90 & 0.76 & 0.80 & 0.88 & 0.87 \\
RoBERTa Chinese (trained by wu3d)         & 0.88 & 0.87 & 0.90 & 0.91 & 0.82 & 0.78 \\
Bert based Chinese (trained by swdd origin) & 0.89 & 0.90 & 0.76 & 0.80 & 0.85 & 0.82 \\
Bert based Chinese (trained by wu3d)        & 0.88 & 0.87 & 0.90 & 0.91 & 0.83 & 0.80 \\
StructBERT-mental (trained by swdd origin) & 0.81 & 0.84 & 0.69 & 0.76 & 0.85 & 0.86 \\
StructBERT-mental (trained by wu3d)         & 0.85 & 0.85 & \textbf{0.92} & \textbf{0.92} & 0.83 & 0.80 \\
Llama3-8b                                     & 0.86 & 0.86 & 0.79 & 0.79 & 0.60 & 0.63 \\
Glm4-9b-chat                                  & 0.76 & 0.81 & 0.77 & 0.81 & 0.79 & 0.82 \\
Qwen2.5 7b                                    & 0.88 & 0.89 & 0.84 & 0.85 & \textbf{0.94} & \textbf{0.94} \\
GPT-4o-Mini                                   & 0.65 & 0.74 & 0.65 & 0.74 & 0.53 & 0.68 \\
GPT-4o                                        & \textbf{0.92} & \textbf{0.92} & 0.91 & 0.92 & 0.83 & 0.86 \\
DeepSeek-R1 671b                              & 0.86 & 0.87 & 0.89 & 0.90 & 0.81 & 0.84 \\
\bottomrule
\end{tabular}
\caption{Model Classification Performance Comparison on Different Test Datasets.}
\label{tab_class_e}
\end{table*}

\begin{figure*}[t]  
    \subsection{Module II Prompt}
    \label{fig:module2}
    \centering
    \includegraphics[width=\textwidth]{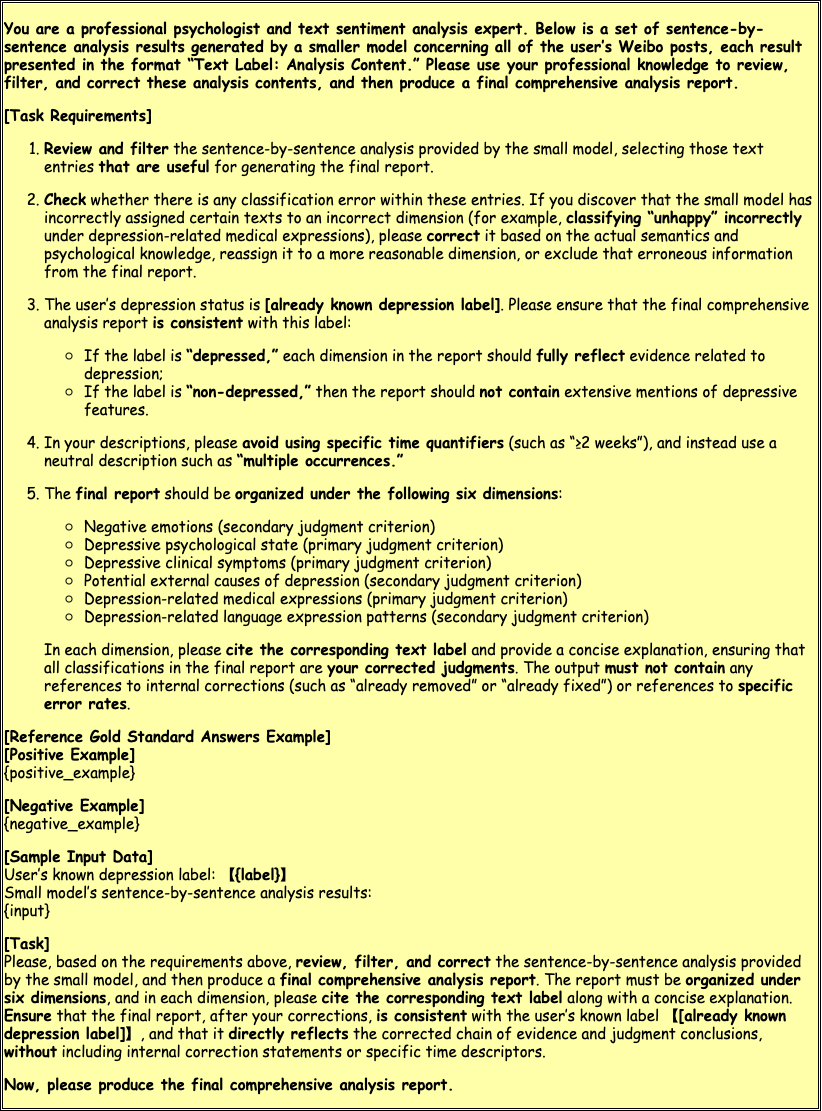} 
    \caption{Module II Prompt.}
\end{figure*}

\end{document}